\documentclass[10pt, a4paper, twocolumn, teaser, showabstract]{naverlabseurope}

\usepackage{lipsum}
\usepackage{multicol}
\usepackage{tikz,tkz-kiviat}
\usepackage{listings}
\usepackage{xcolor}

\newlength{\smallimage}
        \definecolor{rel}{rgb}{.1,.6,.2}
        \definecolor{nrl}{rgb}{1,1,1}
        \definecolor{qim}{rgb}{1,1,1}

\def\egs{\emph{e.g.\,}}
\def\eg{\emph{e.g.}}

\def\ies{\emph{i.e.\,}}
\def\ie{\emph{i.e.}}

\definecolor{lightgray}{gray}{0.93}

\def\be{\begin{equation}}
\def\ee{\end{equation}}
\def\bea{\begin{eqnarray}}
\def\eea{\end{eqnarray}}
\def\ben{\begin{eqnarray*}}
\def\een{\end{eqnarray*}}

\def\bi{\begin{itemize}}
\def\ei{\end{itemize}}

\newcommand{\btab}[1]{\begin{tabular}{#1}}
\newcommand{\etab}{\end{tabular}}
\newcommand{\ba}[1]{\begin{array}{#1}}
\newcommand{\ea}{\end{array}}
\newcommand{\ul}[1]{\underline{#1}}

\def\<{\langle}
\def\>{\rangle}

\newcommand{\bff}{{\bf f}}
\newcommand{\bfg}{{\bf g}}

\newcommand{\bfq}{{\bf q}}

\newcommand{\bfC}{{\bf C}}
\newcommand{\bfD}{{\bf D}}
\newcommand{\bfE}{{\bf E}}
\newcommand{\bfF}{{\bf F}}

\newcommand{\bfM}{{\bf M}}

\newcommand{\bfX}{{\bf X}}

\newcommand{\calG}{{\mathcal G}}

\newcommand{\R}{\mathbb{R}}

\newcommand{\myparagraphnodot}[1]{\vspace{0.1cm}\noindent\textbf{#1}}
\newcommand{\myparagraph}[1]{\myparagraphnodot{#1.}}



\newcommand{\change}[1]{{{#1}}}

\definecolor{DarkGreen}{rgb}{0.5, 0.9, 0.5}

\newcommand{\duster}[0]{DUSt3R}
\newcommand{\panster}[0]{PanSt3R}

\newcommand{\muster}[0]{MUSt3R}
\newcommand{\ludvig}[0]{LUDVIG}
\newcommand{\dino}[0]{DINOv2}
\newcommand{\maskform}[0]{Mask2Former}

\newcommand{\snpp}{ScanNet$\mathbin{++}$}
\newcommand{\snppflat}{ScanNet++}

\newcommand{\PH}{Hypersim}
\newcommand{\PS}{ScanNet}
\newcommand{\PR}{Replica}

\newcommand{\encm}{\textsc{Enc}^{\text{M}}} 
\newcommand{\encd}{\textsc{Enc}^{\text{D}}} 
\newcommand{\dec}{\textsc{Dec}} 
\newcommand{\decp}{\textsc{Dec}^{\text{P}}} 
 
\newcommand{\lin}{\textsc{Lin}} 
 
\newcommand{\head}{\textsc{Head}^{\text{3D}}}

\newcommand{\PQsc}[0]{PQ}

\AtEndPreamble{
    \usepackage[capitalize]{cleveref}
    \crefname{section}{Sec.}{Secs.}
    \Crefname{section}{Section}{Sections}
    \Crefname{table}{Table}{Tables}
    \crefname{table}{Tab.}{Tabs.}
}

\lstnewenvironment{mypython}{
    \lstset{
        language=Python, 
        basicstyle=\footnotesize\ttfamily, 
        keywordstyle=\color{blue}, 
        commentstyle=\color{green!50!black}, 
        frame=single,          
        breaklines=true, 
        breakatwhitespace=true, 
        columns=fullflexible, 
        linewidth=\linewidth,
        rulecolor=\color{gray!90}
    }
}{}
\usepackage[linesnumbered,ruled,vlined]{algorithm2e}
\usepackage{pythonhighlight}

\usepackage{multirow}
\usepackage{bm}
\usepackage{graphicx}
\usepackage{subfig}
\usepackage{amsmath}  
\usepackage{dsfont}   

\title{\panster{}: Multi-view Consistent Panoptic Segmentation}

\correspondingauthor{lojze.zust@naverlabs.com}

\authors{Lojze \v{Z}ust, \authsep 
 Yohann Cabon, \authsep 
Juliette Marrie, \authsep 
Leonid Antsfeld, \authsep 
Boris Chidlovskii, \authsep 
Jerome Revaud   and \authsep 
 Gabriela Csurka }

\affiliations{NAVER LABS Europe} 
\contributions{}
\website{https://europe.naverlabs.com}
 \websiteref{\href{https://europe.naverlabs.com/}}
\teaserfig{\includegraphics[width=0.95\textwidth]{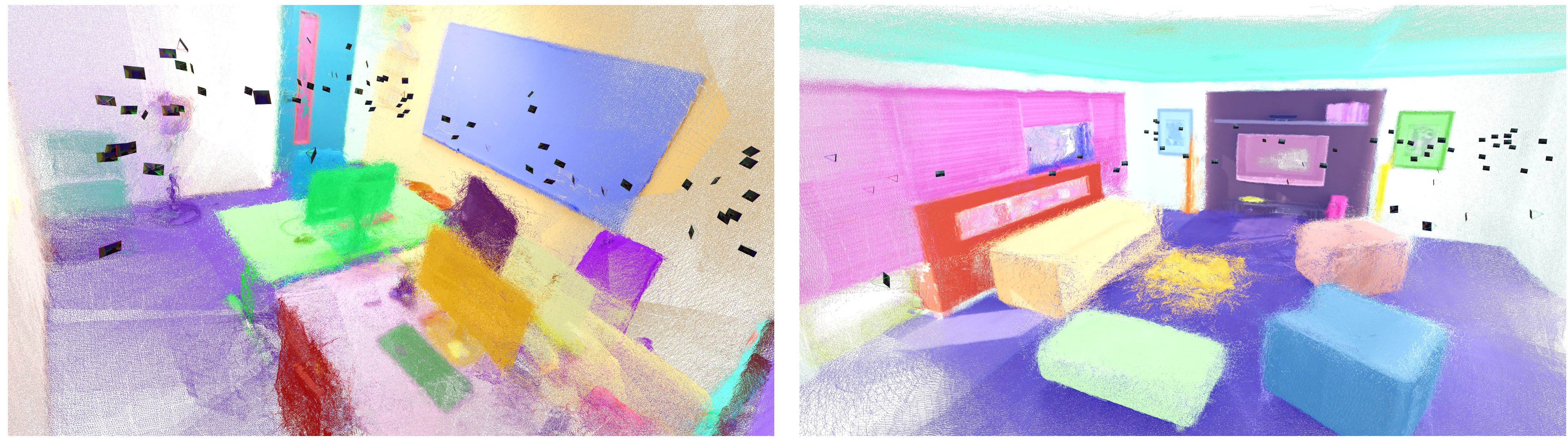}}
\teasercaption{\panster{} jointly predicts 3D geometry and panoptic segmentation of a scene in a single forward pass.  \label{fig:teaser}}
\date{}

\begin{abstract}

Panoptic segmentation of 3D scenes, involving the segmentation and classification of object instances in a dense 3D reconstruction of a scene, is a challenging problem, especially when relying solely on unposed 2D images.
Existing approaches typically leverage off-the-shelf models to extract per-frame 2D panoptic segmentations, before optimizing an implicit geometric representation (often based on NeRF) to integrate and fuse the 2D predictions.
We argue that relying on 2D panoptic segmentation for a problem inherently 3D and multi-view is likely suboptimal as it fails to leverage the full potential of spatial relationships across views.
In addition to requiring camera parameters, these approaches also necessitate computationally expensive test-time optimization for each scene.
Instead, in this work, we propose a unified and integrated approach \panster{}, 
which eliminates the need for test-time optimization by jointly predicting 3D geometry and multi-view panoptic segmentation in a single forward pass.
Our approach builds upon recent advances in 3D reconstruction, specifically upon \muster{}, a scalable multi-view version of \duster{}, 
and enhances it with semantic awareness and multi-view panoptic segmentation capabilities.
We additionally revisit the standard post-processing mask merging procedure and introduce a more principled approach for multi-view segmentation. 
We also introduce a simple method for generating novel-view predictions based on the predictions of \panster{} and vanilla 3DGS.
Overall, the proposed \panster{} is conceptually simple, yet fast and scalable, and achieves state-of-the-art performance on several benchmarks, while being orders of magnitude faster than existing methods. 
\end{abstract}

\begin{document}

\maketitle


\section{Introduction}
\label{sec:intro}

Robust understanding of semantics of 3D scenes is key to many applications like virtual reality, robot navigation or autonomous-driving.
Such use cases require an accurate decomposition of the 3D environment into separate object instances of known classes.
In 2D vision, this joint task of semantic and instance segmentation, denoted as panoptic image segmentation~\cite{KirillovCVPR19PanopticSegmentation}, consists of instance segmentation of \textit{things} classes (\ie countable objects such as cars), 
and semantic segmentation of \textit{stuff} classes (\ie uncountable classes such as road or sky). 
Following \cite{KirillovCVPR19PanopticSegmentation}, a large amount of solutions were proposed for 2D panoptic segmentation, based on CNNs~\citep{XiongCVPR19UPSNetAUnifiedPanopticSegmentationNetwork,LiCVPR19AttentionGuidedUnifiedNetworkPanopticSegm,LiuCVPR19AnEnd2EndNetworkPanSegm,MohanIJCV21EfficientPSEfficientPanopticSegm,ChengCVPR20PanopticDeepLabBottomUpPanopticSegm}, Transformers~\cite{ChengCVPR22MaskedAttentionMaskTransformer4UniversalSiS,DingICML23OpenVocUniversalImgSegmMaskCLIP,XuICCV23MasQCLIP4OpenVocImgSegm,ZouCVPR23GeneralizedDecoding4PixelImageLanguage,QinCVPR23FreeSegUnifiedUniversalOpenVocSegm,LiCVPR24OMGSegIsOneModelGoodEnough4AllSegm}, or more recently diffusion models~\cite{XuCVPR23OpenVocPanopticSegmentationWithText2ImDiffModels,ChenICCV23AGeneralistFramework4PanopticSegm,WangX23DFormerDiffusionGuidedTransformer4UniversalImgSegm,VanGansbekeECCV24ASimpleLatentDiffusion4PanopticSegm}.

A number of recent works~\cite{LiCVPR22PanopticPHNetHighPrecisionLiDARPanopticSegmViaClusteringPseudoHeatmap,HongCVPR21LiDARbasedPanopticSegmDynamicShiftingNetwork,RazaniICCV21GPS3NetGraphbasedPanopticSparseSegm,SirohiTROB21EfficientLPSEfficientLiDARPanopticSegm,XuAAAI22SparseCrossScaleAttentionNetworkLiDARPanSegm} have extended panoptic segmentation to 3D scenes represented as point-clouds, meshes or voxels. 
These methods typically take a 3D representation (\eg a point cloud) as input and label it using neural networks, such as PointNet~\cite{QiCVPR17PointNet3DClassificationSegmentation,qi2017pointnetpp}, designed for direct operation on such data.
However, acquiring dense and accurate point clouds requires dedicated sensors and recent models~\cite{schroppel2022benchmark,WangCVPR24DUSt3RGeometric3DVisionMadeEasy,LeroyECCV24GroundingImageMatchingwithMASt3R,wang2024spanner,tang2024mvdust3rsinglestagescenereconstruction} struggle with noisy or sparse point clouds derived from unposed images.

Instead, in this work, we propose to
\emph{jointly} perform 3D reconstruction and panoptic segmentation given an unconstrained set of unposed images or video frames. 
In this sense our method is closer to NeRF-based~\cite{ZhiICCV21InPlaceSceneLabellingImplicitSceneRepr,Fu3DV22PanopticNeRF3DTo2DLabelTransferSiS,KunduCVPR22PanopticNeuralFieldsSemanticObjecAwareRepr,WangICLR23DMNeRF3DSceneGeometryDecomFrom2D,SiddiquiCVPR23PanopticLiftingFor3DWithNeuralFields,BhalgatNIPS24ContrastiveLift3DObjectInstSegm} or 3D Gaussian Splatting (3DGS)-based methods~\cite{WangX24PLGSRobustPanopticLiftingW3DGaussianSplatting} that start from a collection of 
images. 
These approaches typically rely on posed images and off-the-shelf 2D panoptic segmentation models~\cite{ChengCVPR22MaskedAttentionMaskTransformer4UniversalSiS}, followed by lifting and fusing the input panoptic information in 3D via NeRF~\cite{MildenhallECCV20NeRFRepresentingScenesAsNeuralRadianceFields} or 3DGS~\cite{KerblTOG233DGaussianSplatting4RealTimeRadianceFieldRendering}.

While these methods allow aggregation of potentially inconsistent and noisy 2D panoptic labels from multiple images into consistent 3D labels, they have several limitations: (1) they depend on accurate camera poses, (2) they require costly test-time optimization to align 2D segmentations with 3D geometry, and (3) they inherently separate the 2D segmentation and 3D reconstruction pipelines, potentially sacrificing efficiency and accuracy. 

We argue that 3D reconstruction and 3D panoptic segmentation are two intrinsically connected tasks, both involving reasoning in terms of 3D geometry of the scene and its instance decomposition. Therefore, we propose to model the 3D geometry and its panoptic segmentation in a unified, end-to-end framework that performs directly multi-view consistent panoptic segmentation. Prior works in such panoptic 3D reconstruction usually focus on single-image inputs~\cite{dahnert2021panoptic,zhang2023uni3d} or require posed video inputs~\cite{wu2024panorecon}.

Instead, building on top of the recent 3D reconstruction network \muster{} \cite{WangCVPR24DUSt3RGeometric3DVisionMadeEasy},
we propose
\panster{} (\ul{Pan}optic MU\ul{St3R}) which jointly predicts the 3D scene geometry and its panoptics from an \textit{unconstrained} collection of \textit{unposed} images in a single forward pass. \panster{} leverages two pre-trained feature extractors to encode frames in both semantic (2D) and 3D-aware information, then directly regresses 3D geometry via a 3D head, and performs multi-view instance mask prediction via a \maskform{}-like decoder. 
These mask predictions are finally filtered using a lightweight novel quadratic binary optimization framework (QUBO).
This turns out to be crucial step in our method, as we show that the standard filtering technique is poorly suited for multi-view predictions.
Finally, we demonstrate that it is straightforward to generate novel view panoptic predictions based on the outputs of our method via simple test-time uplifting of labels to 3DGS~\cite{MarrieX24LUDVIG}. 

In summary, our main contributions are as follows.
We introduce a method for joint 3D reconstruction and panoptic segmentation, which 
tackles the problem with a single forward pass. The approach is simple, fast, and operates on hundreds of images
without requiring any camera parameters or test-time optimization.
Second, we propose a novel mask prediction merging framework grounded in a mathematically rigorous formulation, that significantly improves 
the quality of multi-view panoptic predictions.
Third, we explore two distinct approaches
for predicting panoptic segmentation from novel viewpoints within our framework, leveraging simple yet effective methodologies.
Finally, we conduct extensive evaluation 
and ablative studies on several datasets obtaining state-of-the-art results both in terms of panoptic quality and inference speed.

\section{Related work}
\label{sec:related}

\myparagraphnodot{2D Panoptic Segmentation}
is a unification of the semantic and instance segmentation tasks.
Its goal is to decompose an image into different regions, each region corresponding to an individual object (denoted as \textit{thing}) or uncountable concepts like `sky' or `ground' (denoted as \textit{stuff}).
The first panoptic methods extended Mask R-CNN~\citep{HeICCV17MaskRCNN} to design a deformable-convolution-based semantic segmentation head and solve the two sub-tasks simultaneously~\cite{KirillovCVPR19PanopticSegmentation,deGeusX18PanopticSegmJointSiSAndInstSNetwork,XiongCVPR19UPSNetAUnifiedPanopticSegmentationNetwork,MohanIJCV21EfficientPSEfficientPanopticSegm,KirillovCVPR19PanopticFeaturePyramidNetworks}. 
Another set of models \cite{PorziCVPR19SeamlessSceneSegmentation,YangX19DeeperLabSingleShotImageParser,ChengCVPR20PanopticDeepLabBottomUpPanopticSegm,WangECCV20AxialDeepLabStandAloneAxialAttPanopticSegm} build upon the DeepLab architecture~\citep{ChenICLR15SemanticImgSegmFullyConnectedCRFs}. 
Instead,
\citep{LiCVPR19AttentionGuidedUnifiedNetworkPanopticSegm} combines a proposal attention module with a mask attention module, 
\citep{LiuCVPR19AnEnd2EndNetworkPanSegm} propose an end-to-end occlusion-aware pipeline, and \citep{SofiiukICCV19AdaptISAdaptiveInstanceSelectionNetwork} a fully differentiable end-to-end network for class-agnostic instance segmentation jointly trained with a semantic segmentation branch. 
\citet{GaoFuICCV19SSAPSingleShotInstSegmAffinityPyramid} jointly train semantic class labeling with a pixel-pair affinity pyramid and 
\citet{YuanECCV20ObjectContextualReprSemSegm} generalized object-contextual representations to panoptic segmentation.

With the success of Vision Transformers, 
Mask2Former~\citep{ChengCVPR22MaskedAttentionMaskTransformer4UniversalSiS}, inspired by DETR ~\cite{CarionECCV20DETREnd2EndObjectDetectionWithTransformers}, adopted a more unified approach to directly produce panoptic output, posing the task as a mask prediction and classification problem.
Several recent extensions also aim for open-vocabulary segmentation capabilities (\egs using a CLIP text encoder)~\cite{DingICML23OpenVocUniversalImgSegmMaskCLIP,XuICCV23MasQCLIP4OpenVocImgSegm,ChenICCV23OpenVocPanopticSegmWithEmbeddingModulation,YuNIPS23ConvolutionsDieHardOpenVocSegmFrozenConvCLIP,HeCVPR23PrimitiveGenerationSemanticAlignment4UnivZeroShotSegm,LiCVPR24OMGSegIsOneModelGoodEnough4AllSegm,QinCVPR23FreeSegUnifiedUniversalOpenVocSegm,GuNIPS23DaTaSegTamingUnivMultiDatasetMultiTaskSegmModel}. 
Recently, several diffusion based methods were also proposed for this task~\cite{XuCVPR23OpenVocPanopticSegmentationWithText2ImDiffModels,ChenICCV23AGeneralistFramework4PanopticSegm,WangX23DFormerDiffusionGuidedTransformer4UniversalImgSegm,VanGansbekeECCV24ASimpleLatentDiffusion4PanopticSegm}.

\myparagraphnodot{3D Panoptic Segmentation} is a direct extension of 2D panoptic segmentation for 3D scenes.
We can distinguish between several categories of approaches.
First, methods that directly process an input 3D point-cloud, typically obtained by dedicated sensors (ToF or LIDAR), thereby assuming prior knowledge of the 3D scene geometry~\cite{LiCVPR22PanopticPHNetHighPrecisionLiDARPanopticSegmViaClusteringPseudoHeatmap,HongCVPR21LiDARbasedPanopticSegmDynamicShiftingNetwork,RazaniICCV21GPS3NetGraphbasedPanopticSparseSegm,SirohiTROB21EfficientLPSEfficientLiDARPanopticSegm,XuAAAI22SparseCrossScaleAttentionNetworkLiDARPanSegm}.

The second category of methods, closer to our approach, require only a set of input images and respective camera parameters (if not provided directly, the latter is usually obtained via standard SfM techniques~\cite{colmap}). 
Existing approaches in this category are either based on NeRF~\cite{MildenhallECCV20NeRFRepresentingScenesAsNeuralRadianceFields}, or Gaussian Splatting \cite{KerblTOG233DGaussianSplatting4RealTimeRadianceFieldRendering}, with implicit or explicit labeled 3D representations as output, respectively. 
These methods typically perform 3D panoptic segmentation by lifting 2D segmentation masks obtained with pre-trained 2D panoptic segmentation models (\egs Mask2Former \cite{ChengCVPR22MaskedAttentionMaskTransformer4UniversalSiS}) to 3D. 
\citet{ZhiICCV21InPlaceSceneLabellingImplicitSceneRepr} showed that noisy 2D semantic segmentations can be fused into a consistent volumetric model by a NeRF, and their model was extended to instance and panoptic segmentation in \cite{Fu3DV22PanopticNeRF3DTo2DLabelTransferSiS,KunduCVPR22PanopticNeuralFieldsSemanticObjecAwareRepr,WangICLR23DMNeRF3DSceneGeometryDecomFrom2D}. 

Panoptic NeRF~\cite{Fu3DV22PanopticNeRF3DTo2DLabelTransferSiS} starts from a set of sparse images, coarse 3D bounding primitives and noisy 2D predictions to generate panoptic labels via volumetric rendering. Panoptic Neural Fields (PNF)~\cite{KunduCVPR22PanopticNeuralFieldsSemanticObjecAwareRepr} learns a panoptic radiance field with a separate instance MLP and a semantic MLP by explicitly decomposing the scene into a set of objects and amorphous background. These MLPs collectively define the panoptic-radiance field describing 3D point color, semantic and instance labels. 
DM-NeRF 
\cite{WangICLR23DMNeRF3DSceneGeometryDecomFrom2D} 
introduced an object field component to learn unique codes for all individual objects in 3D space from 2D supervision and panoptic segmentation with an extra semantic branch parallel to object code branch. 
Panoptic Lifting (PanLift)~\cite{SiddiquiCVPR23PanopticLiftingFor3DWithNeuralFields} 
relies on TensoRF \cite{ChenECCV22TensoRFTensorialRadianceFields} on top of which they introduce lightweight output heads for learning semantic and instance fields. The core idea of Contrastive Lift
\cite{BhalgatNIPS24ContrastiveLift3DObjectInstSegm}
is a slow-fast clustering objective function well-suited for scenes with a large number of objects.
They lift 2D segments to 3D fusing them by means of a neural field representation, which encourages multi-view consistency across frames.

On the Gaussian Spatting side, 
PLGS \cite{WangX24PLGSRobustPanopticLiftingW3DGaussianSplatting}
learns to embed an additional semantic and instance probability vectors for each Gaussian, which can be rendered on novel views in parallel to RGB. 
To handle noisy panoptic predictions, they rely on Scaffold-GS architecture~\cite{LuCVPR24ScaffoldGSStructured3DGaussians4ViewAdaptiveRendering} where additional depth maps are provided as input and 3D Gaussians are initialized with semantic anchor points used for smooth regularization during training. 
Instead, 
PCF-Lift~\cite{zhu24pcf-lift}
addressed the degradation of performance in complex scenes caused by noisy and error-prone 2D segmentations by introducing Probabilistic Contrastive Fusion (PCF), which learns to embed probabilistic features to robustly handle inaccurate segmentations and inconsistent instance IDs.

Alternatively, a category of methods explores joint prediction of 3D geometry and panoptics of the scene. However, these approaches are either limited to single-image inputs~\cite{dahnert2021panoptic,chu2023buol,zhang2023uni3d}, or rely on posed and ordered collection of input frames~\cite{wu2024panorecon,zhou2024eprecon}.

In contrast to all these methods, our approach works on collections of unordered and unposed input images without camera parameters or depth maps, and directly outputs a 3D reconstruction annotated with panoptic labels in a single forward pass (see 
examples in \cref{fig:teaser}).

\begin{figure*}[ttt]
    \centering
    \includegraphics[width=\linewidth]{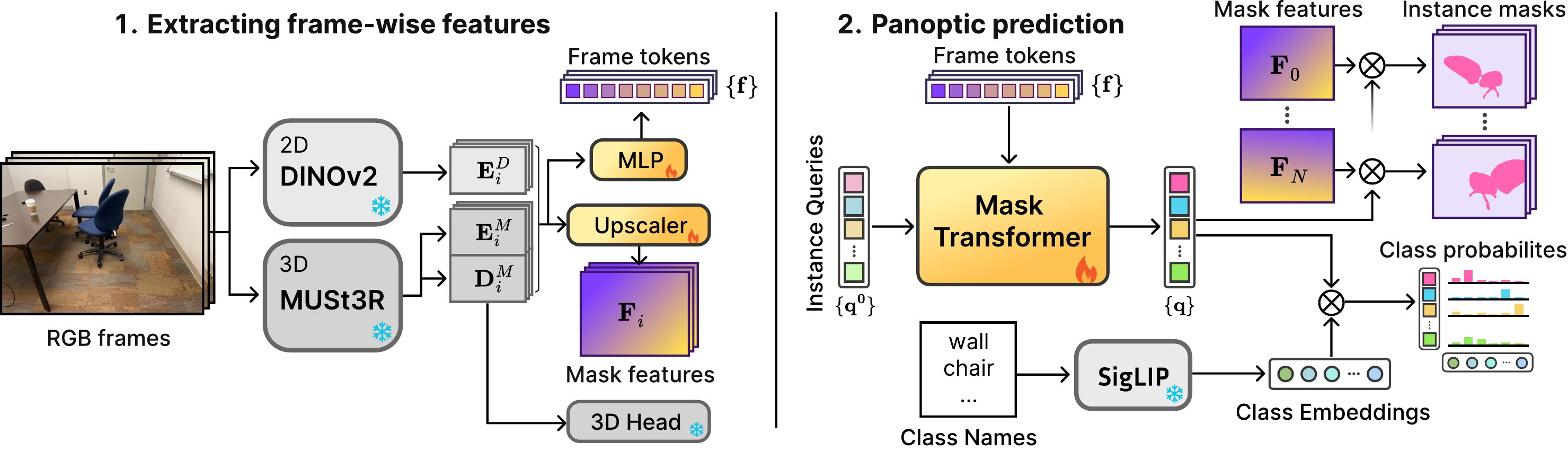}
    \caption{ Architecture of \panster{}. First, the input un-posed RGB frames are passed through pretrained DINOv2 and MUSt3R to extract 2D semantic and globally aligned 3D features respectively.
    Frame tokens and mask features are constructed from per-frame features. Finally, a mask transformer is used to decode instance masks and their class probabilities, by cross-attending learnable instance queries with extracted frame tokens. 
    }
    \label{fig:pipeline}
\end{figure*}

\section{Method}
\label{sec:method}

\myparagraph{Problem statement}
Given a set of $N$ images $I_1 \ldots I_N \in \R^{W\times H \times 3}$, we aim to jointly perform 3D reconstruction and panoptic segmentation, outputting for each pixel of each input image a 3D point, a class and an instance ID.
Formally, these outputs materialize as 3D point-maps $\bfX \in \R^{N\times W \times H \times 3}$, semantic segmentation masks $\bfM^\textsc{cls} \in \{1\ldots C\}^{N\times W \times H}$ and instance segmentation masks $\bfM^\textsc{inst} \in \{1\ldots m\}^{N\times W \times H}$, where $W$ and $H$ denotes the image width and height, $C$ is the number of classes and $m$ is the maximum number of instances\footnote{
Instance IDs are not shared between classes and uniquely identify each object or stuff region in the scene.} in the scene (\egs $m=200$).

\myparagraph{Summary}
Our method builds upon recent progress made in the 3D reconstruction community.
Specifically, our approach is based on \muster{}~\cite{cabon2025must3rmultiviewnetworkstereo}, a Transformer-based powerful and scalable 3D reconstruction method, which we augment with panoptic capabilities inspired by Mask2Former~\cite{ChengNIPS21MaskFormerPerPixelClassifIsNotAllYouNeed4SemSegm}.
In~\cref{sec:panster} we detail the overall architecture of our network.
Since the network outputs raw mask predictions that are possibly overlapping, a merging step is necessary to select a globally optimal set of instances (\cref{sec:postproc}).
In order to generate panoptic segmentations for novel viewpoints, we optionally project the labeled point-cloud into a set of 3D Gaussians (\cref{sec:uplift}).

\subsection{\panster{}}
\label{sec:panster} 
The overall \panster{} architecture is illustrated in \cref{fig:pipeline}.
It consists of a feature extraction step that leverages powerful foundational models for 2D and 3D feature extraction, followed by a generation of instance mask proposals.
We detail each step below.

\myparagraph{Feature extraction}
Our network starts by extracting dense semantic and 3D-aware representations from the set of input images by leveraging two pretrained backbones.
Namely, we extract DINOv2 features for each input image, which have been shown to encode well the scene's semantics in a dense manner~\cite{OquabTMLR24DINOv2LearningRobustVisualFeaturesWithoutSupervision}.
Likewise, we extract \muster{}~\cite{cabon2025must3rmultiviewnetworkstereo} features for each input image. 
\muster{} is a recent extension of \duster{}~\cite{WangCVPR24DUSt3RGeometric3DVisionMadeEasy,LeroyECCV24GroundingImageMatchingwithMASt3R}, a foundation model for 3D vision excelling at reconstructing the geometry of a scene given only sparse views.
In practice, \muster{} processes images sequentially while maintaining an internal memory of previously-seen images, thereby allowing to encode multiview-consistent representations. 
Like DINOv2, \muster{} is a Transformer-based network, but it contains an additional decoder to leverage its internal memory. 
This way, it can encode both local and global scene geometry using its encoder and decoder, respectively.

Formally, we denote by $\bfE_n^D=\encd(I_n)$ the DINOv2 feature maps of image $I_n$ and by $\bfE^M_n=\encm(I_n)$ and $\bfD^M_n=\dec(\bfE^M_n)$ the encoder and decoder feature maps from \muster{}. 
Note that by feature maps, we refer to an array of \emph{tokens}, where each token corresponds to a small $16\times16$ patch in the image, \ies we have in reality $\bfE_n^D, \bfE^M_n,\bfD^M_n$ multi-channel feature maps of size $\frac{W}{16}\times \frac{H}{16}$ and the number of channel corresponding to the respective feature dimensions $d_{ED}=d_{EM}=1024$ and $d_{DM}=768$. 
As shown in \cref{fig:pipeline}, the three token maps are concatenated along the feature dimension and passed through an MLP to form compact joint 3D-semantic token representations $\{\bff_n\}$ with $\bff_n \in \R^{d_t}$, where $d_{f}=768$. The concatenated feature maps are also used to construct high-resolution mask features $\bfF_n \in R^{\frac{W}{2}\times \frac{H}{2} \times d_F}$ used for mask prediction, with $d_F=256$.
For that, 
we perform a series of MLP and $2\times$ upsampling operations to gradually upscale them until we reach the output resolution.

\myparagraph{3D geometry}
We leverage \muster{}'s innate capabilities to reconstruct 3D point-clouds. For every image, \muster{} predicts a global point-cloud in the first image's coordinate frame, a local point-cloud, and a confidence map.
Specifically, given decoder features $\bfD^M_n$, a prediction head $\head$ regresses a 3D coordinates and confidences for each pixel: 
\begin{equation}
\textstyle{\bfX^g_{n},\bfX^l_{n},\bfC_{n} =\head(\bfD^M_n) \in \R^{W \times H \times 3}.}
\end{equation}

\myparagraph{Mask prediction and classification}
We follow \maskform{}~\cite{ChengCVPR22MaskedAttentionMaskTransformer4UniversalSiS} in formulating panoptic segmentation as a binary mask prediction and classification problem. We extend this formulation to the multi-view setting, generating globally consistent masks for each instance, \ie 
 the same instance ID is assigned to a 3D object instance across all views it appears in.

This is achieved by a series of learnable queries denoted by $\{\bfq^0_j\}$, shared by all views and used to represent different instances of \textit{things} and \textit{stuff} classes in the input scene. 
These learnable query features can hence be seen as region proposals. They are input to a mask transformer $\decp$ that attends to multi-view frame tokens $\{\bff_n\}$ using cross-attention.
This results in a set of refined queries $\{\bfq_j\}=\decp(\{\bfq_j^0\},\{\bff_n\})$ which serve as the base for instance classification and mask prediction.

To enable training across multiple datasets with diverse labeling conventions, we adopt an open-vocabulary approach for instance classification. Specifically, the class probabilities for each query are computed as the cosine similarity between the query embeddings and SigLIP-generated text embeddings of class names~\cite{zhai2023sigmoidlosslanguageimage,PengCVPR23OpenScene3DSceneUnderstandingWithOpenVocabularies}, \ie
\begin{equation}
p_{i,j} = \text{sim}(\mathbf{q}_j^\textsc{cls}, \mathbf{t}_i),
\end{equation}
where $\bfq_j^\textsc{cls}=\lin^\textsc{cls}(\bfq_j)$ is the class embedding of the query $\bfq_j$ and $\mathbf{t}_i$ is the text embedding of the open-vocabulary class $i$.
Second, the mask prediction of each query is obtained via dot product with the high-resolution mask features $\bfF_n$.
We denote the resulting instance mask for image $I_n$ and query $j$ as 
\begin{equation}
\label{eq:mask_pred}
\textstyle{\bfM_{j,n} = \text{sigmoid}(\bfF_n \cdot \bfq^\textsc{M}_j) \in \R^{\frac{W}{2}\times \frac{H}{2}}},
\end{equation}
where $\bfq_j^\textsc{M}=\lin^\textsc{M}(\bfq_j)$ is the mask embedding of $\bfq_j$. 

\myparagraph{Training loss}
We follow the training protocol of \maskform{}~\cite{ChengCVPR22MaskedAttentionMaskTransformer4UniversalSiS}, which comprises three losses. The focal loss $\mathcal{L}_{cls}$ serves to learn the instance classification~\cite{li2020generalizedfocalloss}.
We use a combination of dice loss $\mathcal{L}_{dice}$ \cite{diceloss} and binary cross-entropy $\mathcal{L}_{bce}$ to learn mask predictions. The final loss is a weighted combination $\mathcal{L} = \lambda_{c}\mathcal{L}_{cls} + \lambda_{d}\mathcal{L}_{dice} + \lambda_{b}\mathcal{L}_{bce}$.
We set $\lambda_{c}=2, \lambda_{d}=5$ and $\lambda_{b}=5$ as in~\cite{ChengCVPR22MaskedAttentionMaskTransformer4UniversalSiS}.

\myparagraph{Discussion}
While our panoptic prediction network is largely inspired by \maskform{}~\cite{ChengCVPR22MaskedAttentionMaskTransformer4UniversalSiS}, we would like to point out a few key innovations. Most notably, our network is inherently multi-view, processing multiple images simultaneously and directly regressing consistent panoptic segmentations across all frames
This is enabled by leveraging 3D-aware features from \muster{} 
and employing a \emph{shared set of queries}, where each query explicitly targets the same object instance across all view.
Unlike~\cite{ChengCVPR22MaskedAttentionMaskTransformer4UniversalSiS}, we do not construct a multi-resolution feature pyramid, but instead we retain
the original frame tokens in order to limit the memory footprint.
Finally, we adapt the classification head to an open vocabulary enabling training on heterogeneous datasets and improving test-time performance, 
similarly to~\cite{openseed}.

\subsection{Merging mask predictions}
\label{sec:postproc}

Given the set of multi-view mask predictions $\bfM_i \in \mathbb{R}^{N \times \frac{W}{2} \times \frac{H}{2}}$ (\cref{sec:panster}) for each query $i$ our
goal is to find a subset of mask predictions that optimally cover the $N \times \frac{W}{2} \times \frac{H}{2}$ output pixel space
-- minimizing the overlap between selected masks, while reducing the amount of empty regions (\ies holes).
Mathematically, this can be formalized as a quadratic unconstrained binary optimization (QUBO) problem:
$$\textstyle{
    u^* = \max_{u \in \{0,1\}^m} \sum_i u_i Q_i - \sum_{i<j} u_i u_j Q_{i,j},}$$
where $u$ is a boolean assignment of proposals, the weight $Q_i=\sum_{k} \bfM_{i,k}$ represents the weighted area $\left|\mathcal{M}_i\right|$ covered by mask proposal $i$, and $Q_{i,j}$ represents the area in excess when selecting both mask proposals $i$ and $j$ due to their overlap: $$\textstyle{Q_{i,j} = \left|\mathcal{M}_i \cap \mathcal{M}_j\right| = \sum_{k} \min(\bfM_{i,k}, \bfM_{j,k}),}$$
$$ \textrm{since} \,\,\, 
 \, \textstyle{\left|\mathcal{M}_i \cup \mathcal{M}_j\right| = \left|\mathcal{M}_i\right| + \left|\mathcal{M}_j\right| - \left|\mathcal{M}_i \cap \mathcal{M}_j\right|.}$$
To limit the selection of overlapping regions, we further multiply $Q_{i,j}$ by a penalty $\lambda_p>1$, (typically $\lambda_p=2$).
Since QUBO is an NP-hard problem, we rely on simulated annealing~\cite{qubo_simulated_annealing} to find a near-optimal solution efficiently.

With $u^*$ we can construct instance masks $\bfM^\textsc{inst}$ by merging the final assigned masks via per-pixel $\arg\max$. $\bfM^\textsc{cls}$ is finally obtained by assigning the highest probability class $c_j = \arg\max_{i} p_{i,j}$ within the mask area of an instance.

\myparagraph{Discussion}
The standard mask merging procedure, as introduced in MaskFormer~\cite{ChengNIPS21MaskFormerPerPixelClassifIsNotAllYouNeed4SemSegm}, 
is substantially different.
In a nutshell, it consists of first filtering out low-confidence mask predictions to get a pool of candidate masks. This is followed by 
 a pixel-wise voting process to select the most confident mask at each location. Finally, additional filtering is applied to remove predictions that lack sufficient vote support. 
While, this heuristic procedure is simple and typically performs well for single images, it often fails to integrate the multi-view constraints essential for 3D panoptic segmentation. Indeed as shown in \cref{sec:ablations}, the 
QUBO merging strategies 
results in a large boost of performance, thanks to its global optimization of instance masks across all views.

\subsection{Panoptic labels on novel views with 3DGS}
\label{sec:uplift}

In order to compare our model with other methods~\cite{Fu3DV22PanopticNeRF3DTo2DLabelTransferSiS,KunduCVPR22PanopticNeuralFieldsSemanticObjecAwareRepr,WangICLR23DMNeRF3DSceneGeometryDecomFrom2D,SiddiquiCVPR23PanopticLiftingFor3DWithNeuralFields,WangX24PLGSRobustPanopticLiftingW3DGaussianSplatting,BhalgatNIPS24ContrastiveLift3DObjectInstSegm} which evaluate the panoptic performance on unseen views, we additionally rely on Gaussian Splatting (3DGS) \cite{KerblTOG233DGaussianSplatting4RealTimeRadianceFieldRendering}.
We explore two possible strategies: (i) we simply generate novel RGB views with vanilla 3DGS and predict the panoptic segmentation by a simple forward pass of \panster{} on the rendered images; or (ii) we uplift the predicted panoptic segmentations to 3D and  render the segmentations on novel views. 
Since the first strategy is trivial and self-explanatory, we now describe the second strategy in more details.
In the following, we denote 2D images $I_1,\ldots,I_N$ and 
instance mask predictions $\bfM^\textsc{inst}$,
as output by the QUBO mask merging described above.

\myparagraph{Scene optimization}
The 3DGS optimizes the means and covariances of the Gaussian densities, their opacities, and the color function parametrized by spherical harmonics \cite{KerblTOG233DGaussianSplatting4RealTimeRadianceFieldRendering}.
Denoting by $\theta$ the color related parameters and by $\psi$ the other parameters,  
the 3DGS optimizes the following reconstruction loss:
  \begin{equation}
  \label{eq:gs-problem}
  \mathcal{L}_{rgb} = \frac 1N \sum_{n=1}^N \mathcal{L}(I_n, \hat{I}_{n}(\theta,\psi)),
  \end{equation}
where $\hat{I}_{n}(\theta)$ is the rendered frame of the scene in the direction corresponding to view $n$, and $\mathcal L$ is a combination of $\mathcal{L}_1$ and SSIM loss functions~\citep{KerblTOG233DGaussianSplatting4RealTimeRadianceFieldRendering}.

\myparagraph{Panoptic regularization}
Optimizing Gaussians on RGB only may lead to Gaussians spanning across multiple distinct object instances or boundaries of semantic classes, which can negatively impact subsequent label uplifting. To address this, we propose an additional regularization term to align Gaussians to the predicted panoptic masks.

Formally, let $P_n$ be 
the instance segmentation $\bfM^\textsc{inst}_n \in \{1\ldots m\}^{N\times W \times H}$ of image $I_n$, where 
$m$ is corresponding to the number of unique instances.
For the sake of computational efficiency, we label each instance with a unique RGB color
obtaining instance color maps $\tilde{P}_n \in \mathbb{R}^{3 \times W \times H }$ for each image (\ies panoptic image). 
We introduce an additional set of Gaussian color parameters $\hat{\theta}$ and an auxiliary loss during Gaussian optimization supervising the rendering of panoptic colors:
  \begin{equation}  
  \mathcal{L}_{reg}(\theta, \phi) = \frac 1N \sum_{n=1}^N \mathcal{L}_1(\tilde{P}_n, \hat{P}_{n}(\hat{\theta},\psi)),
  \label{eq:GSregterm}
  \end{equation}
where $\hat{P}_{n}(\theta,\phi)$ is the rendered panoptic image.
We optimize the Gaussians with the following 
weighted combination of the two losses:
  \begin{equation}
  \min_{\theta, \phi} \quad \mathcal{L}_{rgb}(\theta,\psi) + \lambda \mathcal{L}_{reg}(\hat{\theta}, \psi)
  \end{equation}
with weight $\lambda$ set to $1$ in all our experiments.  

\myparagraph{Uplifting with \ludvig{}}
To uplift the instance labels into our optimized Gaussian splatting scene, we opt for \ludvig{} \cite{MarrieX24LUDVIG}, a recent 3DGS-based feature uplifting method 
that straightforwardly averages 2D pixel features across all views. Instead of using LUDVIG to uplift features, we utilize it to uplift one-hot encoded instance labels $P^0_n \in \{0,1\}^{m \times W \times H}$ obtained from $P_n$.
We define $\mathcal{S}_i$ as the set of view-pixel pairs $(n,u)$ that a Gaussian $\calG_i$ impacts in the forward rendering process. This impact is quantified by the weight $w_i(n,u)$ resulting from $\alpha$-blending. \ludvig{} defines the 3D feature $\bfg_i$ for the Gaussian $\calG_i$ as the following weighted sum:
\begin{equation}
    \label{eq:uplifting}
    \bfg_i = \sum_{(n,u)\in \mathcal{S}_i }\frac{w_{i}(n, u)}{Z_w} P^0_n(u) 
,~~ {Z_w}=\sum_{(n,p)\in\mathcal{S}_i} w_{i}(n, p),
\end{equation}
After uplifting to 3D and the following reprojection, the final 2D rendered instance label is obtained as the $\arg\max$ along the instance label dimension.


\section{Experimental evaluation}
\label{sec:evaluation}

\subsection{Training details}
\label{sec:setup}

\myparagraph{Training datasets}
To train our method we employ a mix of 2D (single-view) and 3D (set of multi-view posed images) datasets for which ground truth panoptic segmentations are available. Detailed statistics about the training datasets can be found in \cref{tab:datastat}.
\snpp~\cite{ChandanICCV23ScanNetpliusplus3DIndoorScenes} is a dataset comprised of 1006 high resolution 3D indoor scenes with dense semantic (100 semantic class labels) and instance annotations. We use the version V2 of the dataset and follow the official split, \ies 850 scenes for training and 50 scenes for validation.
Aria Synthetic Environments (ASE)~\cite{Avetisyan2024scenescript} is a procedurally-generated synthetic dataset containing 100K unique multi-room interior scenes populated with around 8K 3D objects from which we randomly sampled 750 scenes.
InfiniGen~\cite{RaistrickCVPR24InfinigenIndoorsProceduralGeneration} is another tool to procedurally generate 3D datasets.  We generate 936 indoor scenes of 25 images each. 
Finally, we also leverage two widely used 2D panoptic segmentation datasets, COCO~\cite{LinECCV14MicrosoftCOCO} and ADE20K~\cite{ZhouCVPR17SceneParsingThroughADE20KDataset}, which consist of high-resolution images with precise manual annotations. Adding these datasets is useful to improve generalization and robustness, since they offer a larger visual diversity. To simulate multi-view data on 2D images, we sample several geometric and photometric variants of the input image including crops, rotations, and color jittering.

\myparagraph{Training details}
The DINOv2 and \muster{} backbones (resp. ViT-L, and ViT-L+ViT-B architectures) are initialized with their pretrained weights and frozen during the \panster{} training. In preliminary experiments we observed that fine-tuning \muster{} doesn't impact the final performance much, but incurs significant additional training cost. Since each dataset comes with a different set of classes, we restrict the focal loss supervision $\mathcal{L}_{cls}$ to within the set of ground-truth classes of each dataset during training.

\myparagraph{Test-time keyframes}
Since the number of test views can be large at inference (\egs hundreds), we adopt the same technique as in \cite{cabon2025must3rmultiviewnetworkstereo} to reduce the computational and memory footprint. 
Namely, we efficiently cluster the set of input images using retrieval techniques and select a small set of 50
keyframes using the farthest-point-sampling (FPS) algorithm to maximize coverage. Frame tokens $\{\bff_n\}$ are then only selected from these keyframes, which is enough to generate relevant queries $\{\bfq_j\}$, as shown in~\cref{sec:ablations}.
We then process the remaining views frame-by-frame, only extracting the per-frame features $\bfF_n$ and directly performing mask prediction via \cref{eq:mask_pred}, with the decoded queries $\{\bfq_j\}$ obtained from the keyframes.

\begin{table}[ttt]
    \centering
    \caption{Statistics for datasets used during training (top) and evaluation (bottom). \textit{"MV"} denotes that multi-view data is available for each scene.}
    \label{tab:datastat}
    \resizebox{0.9\linewidth}{!}{
    \begin{tabular}{cl|cc|cc}
    \toprule
    & Dataset  & real & MV & \# scenes & \# classes \\
    \midrule
    \multirow{5}{*}{\rotatebox{90}{Training}} 
    & \textbf{\snpp}~\cite{ChandanICCV23ScanNetpliusplus3DIndoorScenes}  & \checkmark & \checkmark & 855  &  100 \\
    & \textbf{ASE}~\cite{Avetisyan2024scenescript} & x &   \checkmark  & 750 &     44  \\
    & \textbf{Infinigen}~\cite{RaistrickCVPR24InfinigenIndoorsProceduralGeneration}  & x &   \checkmark  & 936 &   76  \\
    & \textbf{COCO}~\cite{LinECCV14MicrosoftCOCO} & \checkmark & x & 118k &  80 \\
    & \textbf{ADE20k}~\cite{ZhouCVPR17SceneParsingThroughADE20KDataset} & \checkmark & x & 20k & 150 \\
    \midrule
    \midrule
    \multirow{4}{*}{\rotatebox{90}{Eval}} 
    & \textbf{\snpp}~\cite{ChandanICCV23ScanNetpliusplus3DIndoorScenes}  & \checkmark & \checkmark & 50 &  100  \\
    & \textbf{ScanNet}~\cite{DaiCVPR17ScanNetRichlyAnnotated3DReconstructionsIndoor} & \checkmark & \checkmark & 12  & 20  \\
    & \textbf{Hypersim}~\cite{RobertsICCV21Hypersim} & x & \checkmark & 6  & 20 \\
    & \textbf{Replica}~\cite{StraubX19Replica} & x & \checkmark & 7   & 20  \\
    \bottomrule
    \end{tabular}
    }
\end{table}

\subsection{Evaluation metrics}
\label{sec:metrics}

\myparagraph{Panoptic Quality (PQ)}
The Panoptic Quality (PQ) score~\cite{KirillovCVPR19PanopticSegmentation} is defined as 
\begin{equation}
\label{eq:PQ}
\textrm{PQ}=\frac{2 \sum_{(p,g) \in TP} \textrm{IoU}(p,g)}{2\lvert   TP \rvert + \lvert FP \rvert + \lvert FN \rvert},
\end{equation}
where $p$ is a predicted instance 
and $g$ is a GT class instance. 
Intuitively, this score averages IoU of matched segments while penalizing segments with wrong matches (False Positives) or without matches (False Negatives). It can be seen as a combination of two terms, segmentation quality  $\textrm{SQ}=1/\lvert TP \rvert \sum_{(p,g) \in TP} \textrm{IoU}(p,g)$, and a recognition quality 
 $\textrm{RQ}=TP/(2\cdot\lvert TP \rvert + \lvert FP \rvert + \lvert FN \rvert )$.

\myparagraph{Extension to 3D scenes}
PQ can be trivially computed at the scene level by pretending that the scene is a concatenation of all images, effectively tying predictions between all images. This metric, coined scene-PQ ($\textrm{PQ}^{sc}$), was first proposed in \cite{SiddiquiCVPR23PanopticLiftingFor3DWithNeuralFields} to evaluate the results of 3D panoptic segmentation. 
As we always use the scene-PQ metric, we omit the upper-script "$^{sc}$" for brevity in the following. To compute the overall results for a dataset, we average the per-scene PQs across all scenes.

\begin{table}[ttt]
  \centering
  \caption{
  We report results for direct predictions of \panster{} on rendered test images (with and without QUBO), as well as results obtained via the simple 3DGS uplifting approach with \ludvig{}. $^\dagger$Timing for building the 3DGS. 
  Note that given access to target view images, \panster{} can make predictions without the need for 3DGS (and camera parameters).
  }
  \label{tab:panolift}
  \resizebox{\linewidth}{!}{
  \begin{tabular}{rc|ccc|c}
  \toprule
\multirow{2}{*}{Method} & Req. & {\small \textbf{Hyper-}} & {\small \textbf{Rep-}} & {\small \textbf{Scan}} & Time\\ 
       & Poses & {\small \textbf{sim}} & {\small \textbf{lica}} & {\small \textbf{Net}} & (min) \\
\hline
\textbf{DM-NeRF}~ 
\cite{WangICLR23DMNeRF3DSceneGeometryDecomFrom2D}                & \checkmark    & 51.6  & 44.1 & 41.7  & $\sim$ 900\\
\textbf{PNF}~\cite{KunduCVPR22PanopticNeuralFieldsSemanticObjecAwareRepr} & \checkmark       & 44.8  & 41.1  & 48.3 & - \\
\textbf{PanLift} \cite{SiddiquiCVPR23PanopticLiftingFor3DWithNeuralFields}    & \checkmark   & 60.1 & 57.9  & 58.9 & $\sim$ 450 \\ 
\textbf{Contrastive Lift} \cite{BhalgatNIPS24ContrastiveLift3DObjectInstSegm} & \checkmark   & 62.3 & 59.1  & 62.3 & $\sim$ 420 \\
\textbf{PLGS} \cite{WangX24PLGSRobustPanopticLiftingW3DGaussianSplatting}  & \checkmark      & 62.4 & 57.8 & 58.7  &  $\sim$ 120\\
\textbf{PCF-Lift} \cite{zhu24pcf-lift}                                 & \checkmark          &   -  &   -  & 63.5  & - \\
\midrule
\textbf{\panster} w/o QUBO & $\dagger$ & 51.6 & 57.3    &  59.5 & $\sim 4$ ($+ 35^\dagger$)   \\
 \textbf{\panster} & $\dagger$ & {56.5} & \bf{62.0}    &  {65.7}   & $\sim 4.5$ ($+ 35^\dagger$) \\
\textbf{\panster} + \textbf{\ludvig} & \checkmark  & \bf{66.3} & {60.6} &  \bf{67.5} & $\sim 40 $ \\
    \bottomrule
\end{tabular}
}
\end{table}

\subsection{Evaluation on the PanLift benchmark}
\label{sec:panolift}

We first evaluate our method on the Panoptic Lifting (PanLift) benchmark~\cite{SiddiquiCVPR23PanopticLiftingFor3DWithNeuralFields}.
It comprises 12 scenes from ScanNet~\cite{DaiCVPR17ScanNetRichlyAnnotated3DReconstructionsIndoor}, 6 scenes from Hypersim~\cite{RobertsICCV21Hypersim} and 7 scenes from Replica~\cite{StraubX19Replica}
(see details in~\cite{SiddiquiCVPR23PanopticLiftingFor3DWithNeuralFields}).
We use the same splits between seen and unseen (novel) views as in \cite{SiddiquiCVPR23PanopticLiftingFor3DWithNeuralFields,BhalgatNIPS24ContrastiveLift3DObjectInstSegm}.

\begin{figure}[ttt]
    \centering
    \includegraphics[width=\linewidth]{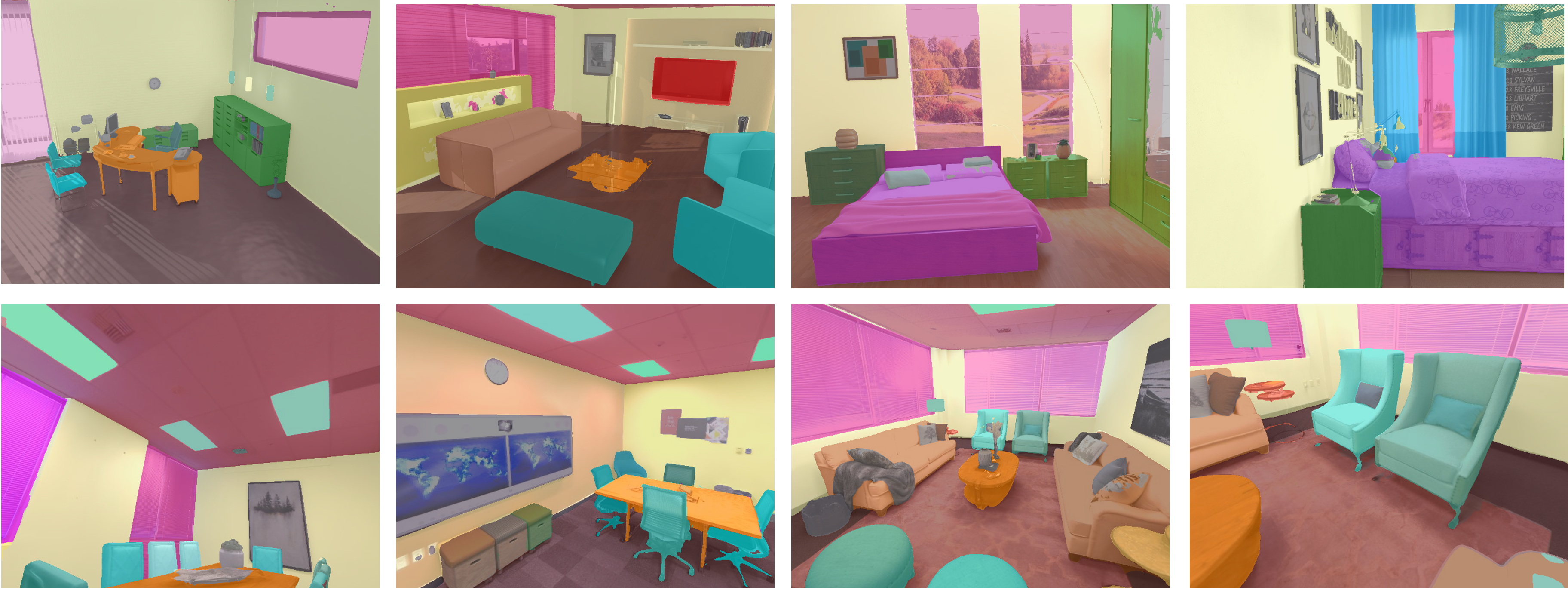}
    \caption{Qualitative examples of novel-view panoptic segmentation on \textbf{Hypersim} and \textbf{Replica} scenes. Predictions are overlaid on top of original images, and colors and their nuances denote different classes and object instances respectively.
    }
    \label{fig:PanoLiftQualitative}
    \vspace{-0.4cm}
\end{figure}

For \panster{}, we experiment with the two strategies presented in \cref{sec:uplift}: (i) we simply render a novel image of the target viewpoint using an off-the-shelf 3DGS model, and perform a forward pass on rendered images with \panster{}; or (ii) we utilize the \ludvig{} uplifting strategy to directly construct a panoptic 3DGS-based scene representation. 
Given access to target-view images, \panster{} can directly predict panoptic segmentation in a forward pass, in principle not requiring poses or 3DGS.
However, to be more inline with the competing methods, which predict panoptic segmentation based on test poses, not images,
we compute the PQ results for \panster{} (and \panster{} w/o QUBO) on test images rendered with vanilla 3DGS built with posed training images\footnote{Note that prediction quality of \panster{} is limited by the fidelity of 3DGS rendered views. Direct prediction on test images yields notably better results (App. \cref{tab:resolution}).}. For \panster{} + \ludvig{} we use the panoptic segmentations of \panster{}, predicted on these training images, to train 3DGS with the regularization term \cref{eq:GSregterm} and to uplift the panoptic segmentation labels into the 3DGS (\cref{sec:uplift}). The panoptic segmentations on test views are then rendered from the uplifted 3DGS features (one-hot instance labels).

\myparagraph{Comparison with existing methods}
3D panoptic segmentation methods either directly process an input 3D point-cloud (see discussion in \cref{sec:SGIFormer}) or they
revolve around the idea of performing
3D panoptic segmentation by lifting 2D segmentation
masks obtained with pre-trained 2D panoptic segmentation models. The latter
leverage off-the shelf 2D panoptic segmentation models, mostly \maskform{}~\cite{WangX24PLGSRobustPanopticLiftingW3DGaussianSplatting} pretrained on COCO~\cite{LinECCV14MicrosoftCOCO} and map the COCO panoptic classes to the following 21 classes: \textit{\textcolor[HTML]{eeea14}{wall}, \textcolor[HTML]{623636}{floor}, \textcolor[HTML]{389138}{cabinet}, \textcolor[HTML]{800080}{bed}, \textcolor[HTML]{00FFFF}{chair}, \textcolor[HTML]{AB6545}{sofa}, \textcolor[HTML]{FF7F00}{table}, \textcolor[HTML]{000f73}{door}, \textcolor[HTML]{EE3196}{window}, 
\textcolor[HTML]{48D6EC}{counter}, 
\textcolor[HTML]{BE7CBE}{shelves}, 
\textcolor[HTML]{00A6FF}{curtain}, 
\textcolor[HTML]{A73B49}{ceiling}, \textcolor[HTML]{72C471}{refrigerator}, \textcolor[HTML]{F11D05}{television}, 
\textcolor[HTML]{2FFF00}{person}, 
\textcolor[HTML]{EE4A4A}{toilet}, 
\textcolor[HTML]{BA92E1}{sink}, 
\textcolor[HTML]{3C9D81}{lamp}, 
\textcolor[HTML]{0000FF}{bag}} and  \textit{\textcolor[HTML]{7A7A7D}{other}}.
In general these methods lift and align these 2D predictions to 3D with a test-time optimization of a NeRF or 3DGS, which is computationally costly and requires information about camera poses.

\begin{figure}[ttt]
    \centering
    \includegraphics[width=\linewidth]{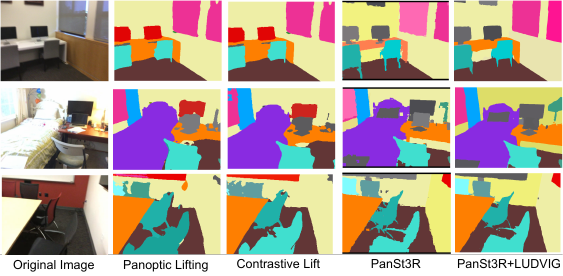}
    \caption{
    Qualitative comparison of novel-view panoptic segmentation on \textbf{ScanNet}~\cite{DaiCVPR17ScanNetRichlyAnnotated3DReconstructionsIndoor} scenes. Colors and their nuances denote different classes and object instances respectively.}
    \label{fig:ComparPanoLift}
        \vspace{-0.4cm}
\end{figure}

We compare our \panster{} variants on the PanoLift datasets 
against state-of-the art methods in \cref{tab:panolift}, where
DM-NeRF~\cite{WangICLR23DMNeRF3DSceneGeometryDecomFrom2D}, PNF~\cite{KunduCVPR22PanopticNeuralFieldsSemanticObjecAwareRepr}, 
PanLift~\cite{SiddiquiCVPR23PanopticLiftingFor3DWithNeuralFields} and Contrastive Lift~\cite{BhalgatNIPS24ContrastiveLift3DObjectInstSegm} are NeRF-based approaches, and PLGS~\cite{WangX24PLGSRobustPanopticLiftingW3DGaussianSplatting} and PCF-Lift~\cite{zhu24pcf-lift}      
rely on 3DGS to uplift 2D panoptic segmentation masks. 
\panster{} inference is performed using the full training class set, then re-mapped to the target 21 classes, similar to previous works.
In \cref{fig:PanoLiftQualitative} we provide visual examples for \panster+\ludvig{} on different datasets and scenes, and in
\cref{fig:ComparPanoLift} we provide qualitative comparisons between \panster{}, \panster+\ludvig{}, PanLift and Contrastive Lift.

\myparagraph{Discussion}
\panster{} + \ludvig{} set a new state-of-the-art, except on Replica, where direct forward pass prediction of \panster{} on rendered views performs slightly better.
On Hypersim and ScanNet, uplifting the panoptic segmentations performs much better than direct forward with \panster{} due to the noise reduction effect of multi-view feature aggregation (\cref{fig:ComparPanoLift}).
Notably, \panster{} accomplishes this while being far more computationally efficient than previous methods, even when uplifting to 3DGS via \ludvig{}.
Additionally, using a simple direct prediction with \panster{} on renedered images is enough to outperform all previous methods on two out of three datasets (except Hypersim),
with potentially no need for camera parameters in contrast to existing methods.

\subsection{Evaluation on \snppflat} 
\label{sec:scannetpp}

We also evaluate \panster{} on the validation set of the \snpp~\cite{ChandanICCV23ScanNetpliusplus3DIndoorScenes}.
For each of the 50 validation scenes, we randomly select 100 frames (only iPhone images) and use \panster{} to predict multi-view consistent panoptic segmentations for these images in a single forward pass. 
We then randomly select 50 images from the remaining pool of images to serve as test views in order to evaluate the panoptic segmentation on novel unseen viewpoints with the same process as used in the PanLift benchmark. A major difference compared to PanLift is a much larger number of classes (100 instead of 20),
including small objects (\egs \textit{crate, paper, socket, cup, smoke detector, soap dispenser}), hence requiring much more fine-grained segmentation. To better assess the performance of the models, we also report PQ$_\text{th}$ and PQ$_\text{st}$, denoting panoptic quality computed separately on \textit{thing} and on \textit{stuff} classes.

\begin{table}[ttt]
  \centering
  \caption{
  Results on the \snpp{} val set. We report result of our default model \panster{} (full) and one trained only on \snpp{}. \panster{} is compared with PanLift~\cite{SiddiquiCVPR23PanopticLiftingFor3DWithNeuralFields} and Contrastive Lift~\cite{BhalgatNIPS24ContrastiveLift3DObjectInstSegm}, utilizing Mask2Former~\cite{ChengCVPR22MaskedAttentionMaskTransformer4UniversalSiS} finetuned on \snpp{}.}
  \label{tab:scannetpp}
    \resizebox{0.9\linewidth}{!}{
 \begin{tabular}{l|ccc|c}
  \toprule
  Method & PQ & PQ$_\text{th}$ & PQ$_\text{st}$ & Time (min)\\
  \midrule
\textbf{PanLift}~\cite{SiddiquiCVPR23PanopticLiftingFor3DWithNeuralFields} 
                               & 29.5 & \change{15.6} & \change{59.4} & $\sim$ 500 \\
\textbf{Contrastive Lift}~\cite{BhalgatNIPS24ContrastiveLift3DObjectInstSegm} 
                               & 28.4 & 14.8 & 56.3 & $\sim$ 460 \\
\midrule
\textbf{\panster}  (\snpp) & 46.7  &   43.2 & 55.8    & $\sim$ 2.3 \\ 
+ \textbf{\ludvig} & 54.8 & 52.4 & 62.4 &$\sim$ 35    \\
\hline
\textbf{\panster} (full)    & 49.1 &  45.8  & 58.7 &  $\sim$ 2.3\\
+ \textbf{\ludvig}  & 54.7 &  51.7  & 62.4   & $\sim$ 35 \\
  \bottomrule
\end{tabular}
}
\end{table}

\begin{figure*}[ttt]
    \centering
    \includegraphics[width=\linewidth]{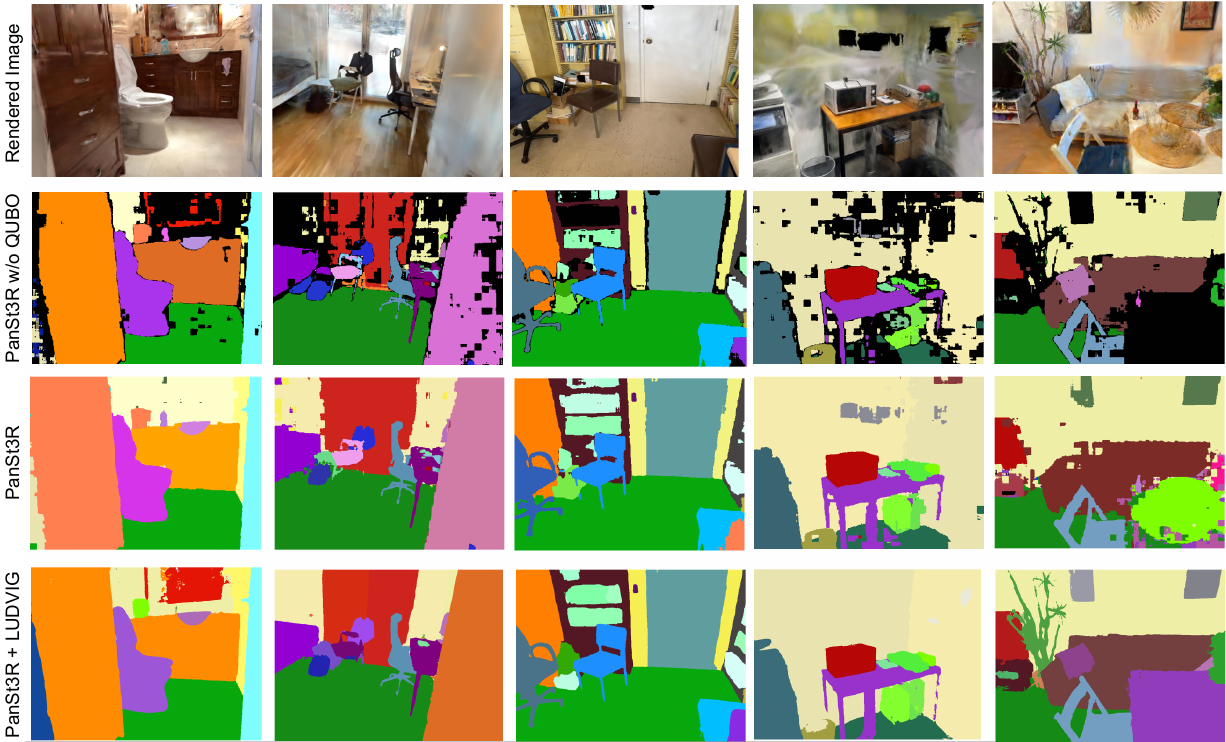}
    \caption{
    Qualitative results of novel-view panoptic segmentation on various \textbf{\snpp}~\cite{DaiCVPR17ScanNetRichlyAnnotated3DReconstructionsIndoor} scenes.  
    Colors and their nuances denote different classes and their instances. Note that due to the large number of classes (100), some may share similar colors.}
    \label{fig:scannetpp}
\end{figure*}

\myparagraph{Comparison with existing methods}

Due to a lack of published results for panoptic segmentation on \snpp, 
we compare our method to PanLift~\cite{SiddiquiCVPR23PanopticLiftingFor3DWithNeuralFields} and Contrastive Lift~\cite{BhalgatNIPS24ContrastiveLift3DObjectInstSegm} using the official code and uplift the predictions of \maskform{}, finetuned on the \snpp training set.
To ensure a fair comparison, we also evaluate a variant of \panster{} trained only on the \snpp{} training set, denoted as \panster{}(\snpp) alongside  
our full model\footnote{The same model weights as in in \cref{tab:panolift}, but using the 100 \snpp{} classes during inference.}. 
Quantitative and qualitative results are presented in \cref{tab:scannetpp} and \cref{fig:scannetpp} respectively.

\myparagraphnodot{Discussion}
We observe that both \panster{} and \panster{} (\snpp) with or without \ludvig{} largely outperform both PanLift and Contrastive Lift by more than 10\% in PQ, while being an order of magnitude faster.
Analysis reveals, that related approaches encounter difficulties with \textit{thing} instances, especially small objects (see PQ$_\text{th}$ scores).
Additionally, we observe, that NeRF optimization struggles when using 'only' 100 training views.
In comparison, \panster{}'s performance is relatively invariant to the number of views, since it builds upon the sparse view reconstruction framework of \duster{} and \muster{} (see ablations in \cref{sec:ablations}).
When we compare \panster{} and \panster{} (\snpp) we observe that the model trained on more data is slightly better, but the gap disappears when using \ludvig{}.
Finally, we once again observe, that uplifting labels with \ludvig{} results in a significant improvement both quantitatively and qualitatively (see \cref{fig:scannetpp}).

\subsection{Ablative studies}
\label{sec:ablations}

We perform a range of comparative experiments and ablative studies to evaluate the impact of various components and model configurations on the method's performance. We primarily assess the role of using 2D \dino{} and 3D \muster{} encoders, as well as the impact of QUBO and GS regularization. Further ablative experiences can be found in the appendix (\cref{sec:futherAblations}).

\myparagraph{Impact of 2D \& 3D features}
We ablate the contribution of our two feature extraction backbones on the final performance.
For this study, we use a smaller architecture, starting from \muster{}/\dino{} with a 224x224 input resolution and trained on the \snpp{} training set only. Result obtained using only the 2D (\dino) or 3D (\muster) features are presented in \cref{tab:abl_feats}.
We observe a clear complementarity effect between 2D semantic features of \dino{} and 3D geometric features of \muster{}.
Without 2D features, the overall \PQsc{} drops by 4\% while without 3D semantic features, it drops by 14.7\%, demonstrating the importance of these two backbones for producing high quality 3D panoptic segmentations. 

\begin{table}[ttt]
  \centering
  \caption{Ablating the importance of 3D (MUSt3R~\cite{cabon2025must3rmultiviewnetworkstereo}) and 2D (DINOv2~\cite{OquabTMLR24DINOv2LearningRobustVisualFeaturesWithoutSupervision}) features on the \snpp{} validation set. Results are shown for  \panster{} (224, \snpp)+\ludvig{}.}
  \label{tab:abl_feats}
  {
  \begin{tabular}{c|ccc}
    \toprule
    features &  PQ & PQ$_\text{th}$ & PQ$_\text{st}$ \\ 
    \midrule
    3D + 2D & 50.4 & 45.4 & 61.1 \\
    3D & 46.4 & 40.7 & 58.8 \\
    2D & 35.7 & 28.4 & 51.9 \\
    \bottomrule
    \end{tabular}
  }
\end{table}

\myparagraph{Mask merging strategy}
We evaluate the impact of the QUBO-based mask merging strategy (\cref{sec:postproc}).
In \cref{tab:panolift} and \cref{fig:scannetpp}, we compare two versions of \panster{}, with and without QUBO (\ies using the standard merging strategy from MaskFormer~\cite{ChengNIPS21MaskFormerPerPixelClassifIsNotAllYouNeed4SemSegm}).
The same experiments are performed with additional \ludvig{} uplifting in \cref{tab:3dscannet_qubo}.
We observe an overall large gap in terms of \PQsc{} (except for \panster+\ludvig{} on Replica), which highlights the inadequacy of the standard merging scheme when dealing with multi-view segmentation. In a sense, this is expected, as the standard scheme is purely heuristic and instances are selected only locally, without considering any global consistency.

\begin{table}[ttt]
  \centering
  \caption{
Analyzing the effect of the QUBO merging strategy (\cref{sec:postproc}) and panoptic regularization for 3DGS (\cref{sec:uplift}). Results are reported for \panster{}+LUDVIG.
  }
  \label{tab:3dscannet_qubo}
    \resizebox{\linewidth}{!}{
 \begin{tabular}{cc|ccc|c}
  \toprule
 Reg & QUBO  & {\PH} & {\PR} & {\PS} & \snpp \\
\midrule
x & x & 58.1 &  60.8  & 60.2 & 50.8 \\
x & \checkmark & 66.7 & 60.7 & 67.3  & 52.0 \\
\checkmark & x & 59.3  & 61.2 & 60.6 & 55.2\\
\checkmark & \checkmark &  
66.3 &   60.6 &  67.5  & 54.7\\
\bottomrule
\end{tabular}
}
\end{table}

\myparagraph{\ludvig{} regularization}

\begin{figure}[ttt]
    \centering
    \includegraphics[width=\linewidth]{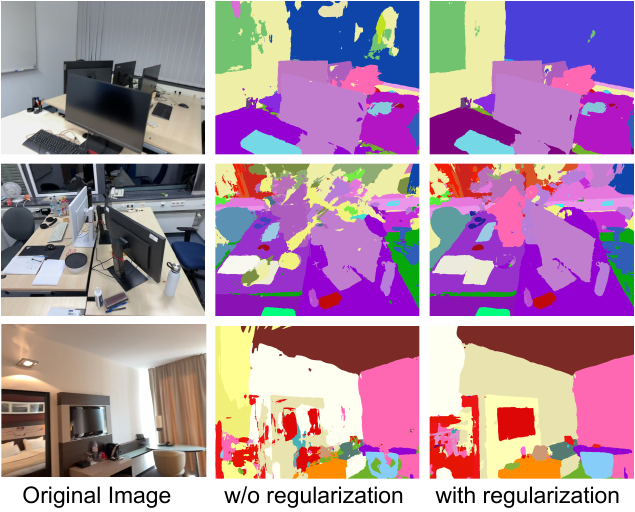}
    \caption{Visual comparison of the effects of panoptic 3DGS regularization on \snpp{}. 
    Results are shown for \panster+\ludvig{} using the standard masking strategy (w/o QUBO).}
    \label{fig:regularization}
\end{figure}

We study the importance of using regularization during the 3DGS optimization (loss $\mathcal{L}_{reg}$ in \cref{sec:uplift}), before the uplifting labels with \ludvig{}. As shown in \cref{tab:3dscannet_qubo}, the regularization helps in most cases (more for the standard mask merging strategy than for QUBO), but not always. Considering visual examples (\cref{fig:regularization}), we observe regularization is more useful in the presence of small objects and visually homogeneous regions (\egs walls), where regularization prevents Gaussians from spanning over the boundary of semantic classes or object instances (\egs wall to ceiling).

\section{Conclusion}
\label{sec:conclusion}

We have presented \panster{}, a novel approach for joint 3D reconstruction and 3D panoptic segmentation operating on unposed and uncalibrated collections of images.
The proposed approach, building upon recent progress in 2D and 3D foundation models, is conceptually simple yet effective, achieving state-of-the-art results on multiple benchmarks.
Despite not relying on any depth input nor camera parameters, and without the need for costly test-time optimization, \panster{} is able to decompose a scene into a set of instances in an efficient manner producing high-quality results and paving the way to promising future applications in the fields of robotics, virtual reality and autonomous driving.

{
    \small
    \bibliographystyle{ieeenat_fullname}
     \bibliography{strings,SemDust3r}
}

\appendix
\vspace{1cm}
\textbf{\Large Appendix }

\section{Further ablative studies}
\label{sec:futherAblations}

\myparagraph{Number of keyframes}
As mentioned in \cref{sec:setup}, \muster{} selects a subset of images called keyframes to reduce the computational and memory load at inference time. In~\cref{tab:3dscannet_kf}, we study the impact of the number of keyframes selected. To better assess the influence of this parameter we evaluate PQ directly on the output of \panster{} on the "seen" set without \ludvig{} uplifting. As expected, the performance significantly decreases when there are too few keyframes. Conversely, we observe the performance plateauing beyond 50 keyframes, meaning that additional information becomes mostly redundant at that point for the scale of scenes in \snpp{}.

\begin{table}[h] 
  \centering
  \caption{Varying the number of keyframes on \snpp{} dataset.  
  }
  \label{tab:3dscannet_kf}
  {
  \begin{tabular}{l|cccccc}
    \toprule
    \# Keyframes &  10 & 20 & 30 & 50 & 100 \\ 
    \midrule
    \textbf{\panster} & 49.3 & 55.0 & 56.3 & 58.2 & 58.2	\\
    \bottomrule
    \end{tabular}
  }
\end{table}

\myparagraph{Number of "seen" images for building NeRF/GS}
In~\cref{tab:3dscannet_nimages} we ablate the effect of reducing the number of images available to build the GS/Nerf and to uplift the panoptic segmentations, to 10, 25, 50 or 100 images. 
PanoLift~\cite{PengCVPR23OpenScene3DSceneUnderstandingWithOpenVocabularies} suffers already when the number of "seen" images is reduced to 50, while \panster{}, both with and without \ludvig, is less sensitive to shrinking the set of "seen" images, its performance dropping only when very few images (10) are available.

\begin{table}[h]
  \centering
  \caption{Varying the number of "seen" images for building GS/NeRF on \snpp{} dataset.
  }
  \label{tab:3dscannet_nimages}
  {
  \begin{tabular}{l|ccccc}
    \toprule
    \# Number of images                  &  10  & 25   & 50    & 100 \\ 
    \midrule
    \textbf{PanoLift~\cite{PengCVPR23OpenScene3DSceneUnderstandingWithOpenVocabularies}}
                                         & 3.3  & 20.2 & 30.6 & 38.5 \\
    \textbf{\panster}                    & 28.4 & 51.8 & 55.8 & 61.8 \\
    \textbf{\panster} + \textbf{\ludvig} & 41.7 & 56.1 & 61.6 & 65.0 \\
    \bottomrule
    \end{tabular}
  }
\end{table}

\myparagraph{Image/model resolution}
We evaluate the impact of the model size (image resolution) in \cref{tab:resolution}. We consider \panster{} trained only on \snpp{} denoted by \panster{(\snpp) and we train either a larger model starting from \muster$_{512}$ or a smaller model starting from \muster$_{224}$. 
We compare the results with and without \ludvig{}. We used the QUBO mask merging strategy in these experiments. 

Note that the results on \panster$_{224}$ are particularly low due to its limitation of square input aspect ratios. Cropping is thus required for this model resulting in missing predictions on the boundary. \ludvig is able to alleviate this issue via the multi-view consolidation of predictions (see examples in \cref{fig:model244}).

\begin{table}[ttt]
  \centering
  \caption{Results of \panster{} trained and evaluated on \snpp. We compare different model resolutions (224 and 512), and different prediction strategies: 
  (i) directly inferring the panoptic segmentation on original test images (orig), (ii) inferring panoptic segmentation on test-view images rendered with vanilla 3DGS (rendered), and (iii) using \ludvig{} to uplift \panster{} predictions. 
  $^\dagger$Note that these results are affected by the square aspect ratio of the 224 model, resulting in missing predictions on the borders (see \cref{fig:model244}). 
  }
  \label{tab:resolution}
    \resizebox{0.9\linewidth}{!}{
 \begin{tabular}{lc|ccc}
  \toprule
  Method & res & PQ & PQ$_\text{th}$ & PQ$_\text{st}$ \\
  \midrule
\textbf{\panster}$^\dagger$ (orig)     & 224 & 39.0 & 36.4 & 48.2 \\
\textbf{\panster}$^\dagger$ (rendered) & 224 & 32.5 & 29.2 & 41.5   \\
\textbf{\panster} + \textbf{\ludvig}   & 224 & 50.4 & 45.4 & 61.1\\
\hline
\textbf{\panster} (orig)      & 512    & 57.3  & 51.7 & 70.4  \\
\textbf{\panster} (rendered)  & 512    &  46.7 & 43.2 & 55.8\\
\textbf{\panster} + \textbf{\ludvig}  & 512 & 54.8 & 52.4 & 62.4  \\
  \bottomrule
\end{tabular}
}
\end{table}

\begin{figure}[ttt]
    \centering
    \includegraphics[width=\linewidth]{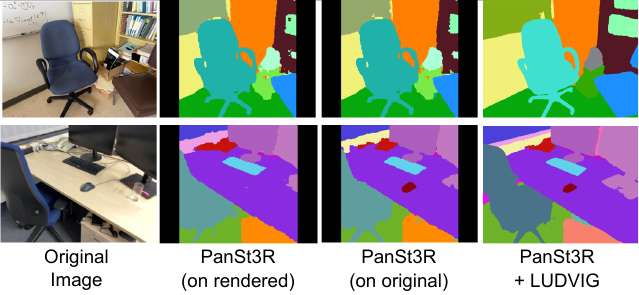}
    \caption{Qualitative panoptic segmentation results with the \panster(244) model.}
    \label{fig:model244}
\end{figure}

\section{Comparison with a two-stage approach}
\label{sec:SGIFormer}

We compare our single-feed forward approach with a two-stage approach of first predicting a point-cloud and running an of the shelf 3D point-cloud panoptic segmentation method to label it.
For this, we considered SGIFormer \cite{yao2025sgiformer}, the top-performing method with available code from the 3D Instance Segmentation Benchmark Leaderboard\footnote{\url{https://kaldir.vc.in.tum.de/scannetpp/benchmark/insseg}}. To compare with \panster's labeling of the point-cloud, SGIFormer is run on the output of 3D MUSt3R head (reconstructed 3D point cloud), and directly on the 3D GT for reference. To evaluate w.r.t. the ground-truth, the labels of the predicted point-cloud are remapped to the GT point-cloud via nearest-neighbor sampling. Results are reported in \cref{tab:3DPCsegm} and qualitative comparison are shown in \cref{fig:3DPCsegm}. We observe that SGIFormer (and similar methods that learn to directly segment the point cloud) works well on the clean ground-truth point-clouds but is very sensitive to noise common in predicted 3D point clouds.

\begin{table}[hhh]
\centering
\caption{Point-cloud instance segmentation on ScanNet++V2.}
\label{tab:3DPCsegm}
\resizebox{\linewidth}{!}{
\begin{tabular}{@{}lcccc@{}}
\toprule
\textbf{Method}       & \textbf{AP50} & \textbf{AP25} & \textbf{Pr} & \textbf{Re} \\
\midrule
MUSt3R pcd + SGIFormer~\cite{yao2025sgiformer}    & 1.8           & 3.0           & 27.5        & 2.5         \\
PanSt3R               & 19.1          & 32.9          & 45.7        & 22.9        \\
GT pcd + SGIFormer~\cite{yao2025sgiformer}       & 33.2          & 40.5          & 54.0        & 39.8        \\
\bottomrule
\end{tabular}}
\end{table}

\begin{figure}[t]
    \centering
    \includegraphics[width=\linewidth]{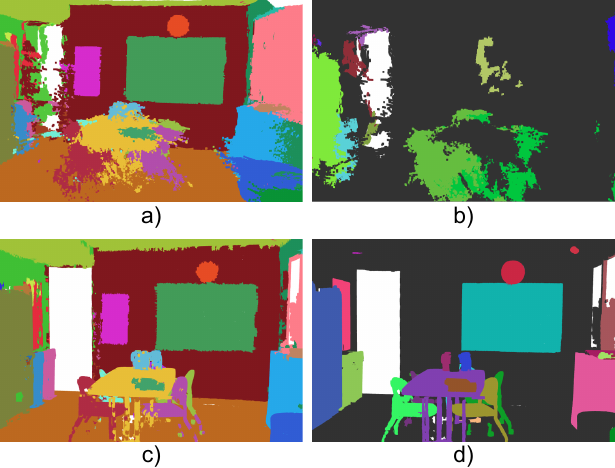}
    \caption{Point-cloud labeling: (a) output of PanSt3R, (b) SGIFormer prediction on MUSt3R point-cloud,  
    (c) PanSt3R output remapped to GT point-cloud, (d) SGIFormer on GT point-cloud. (Instance colors are not aligned between \panster{} and SGIFormer)
    }
    \label{fig:3DPCsegm}
\end{figure}

\begin{figure}[ttt]
    \centering
    \includegraphics[width=0.9\linewidth] {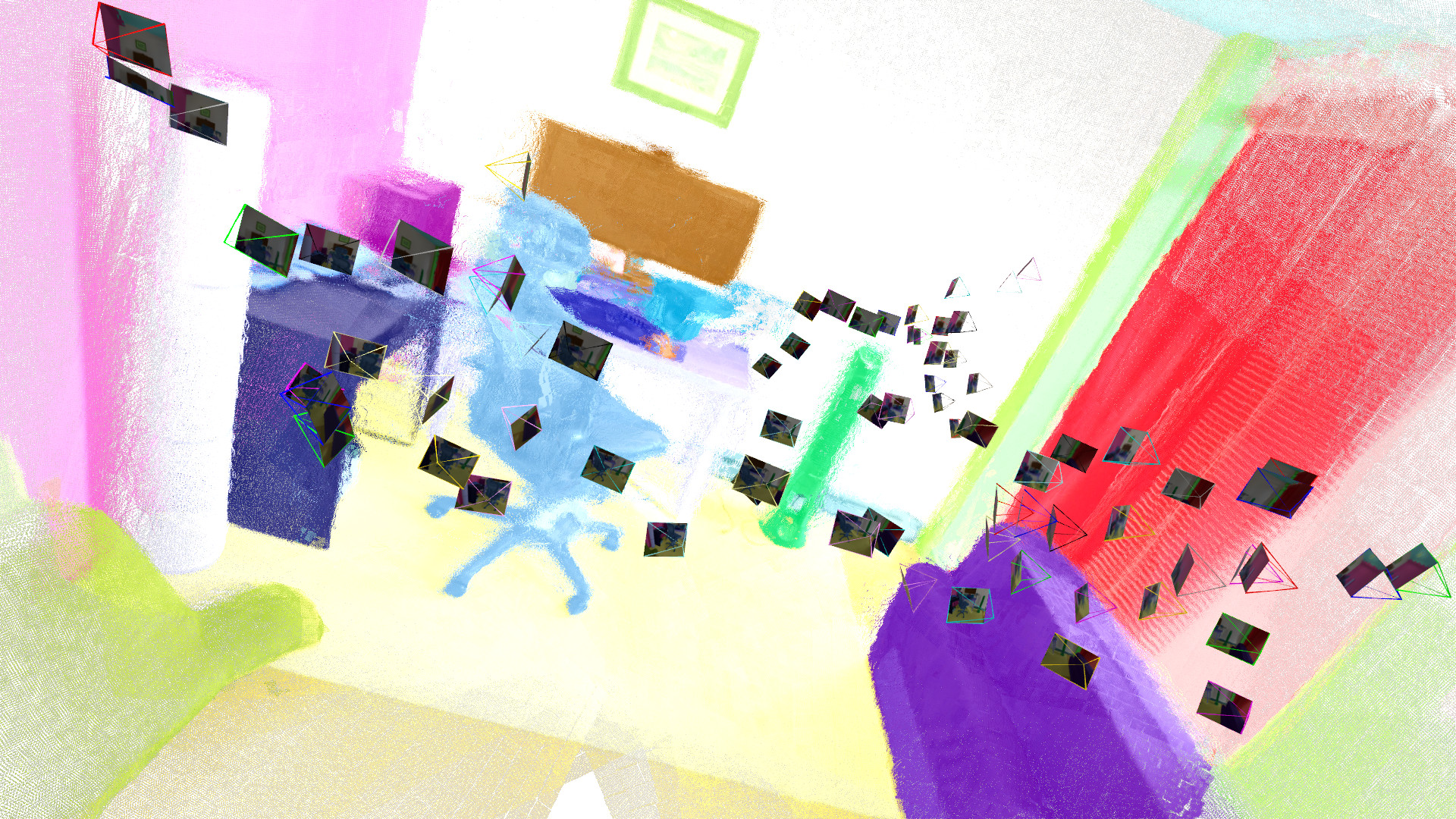}\\ \, \\
     \includegraphics[width=0.9\linewidth]{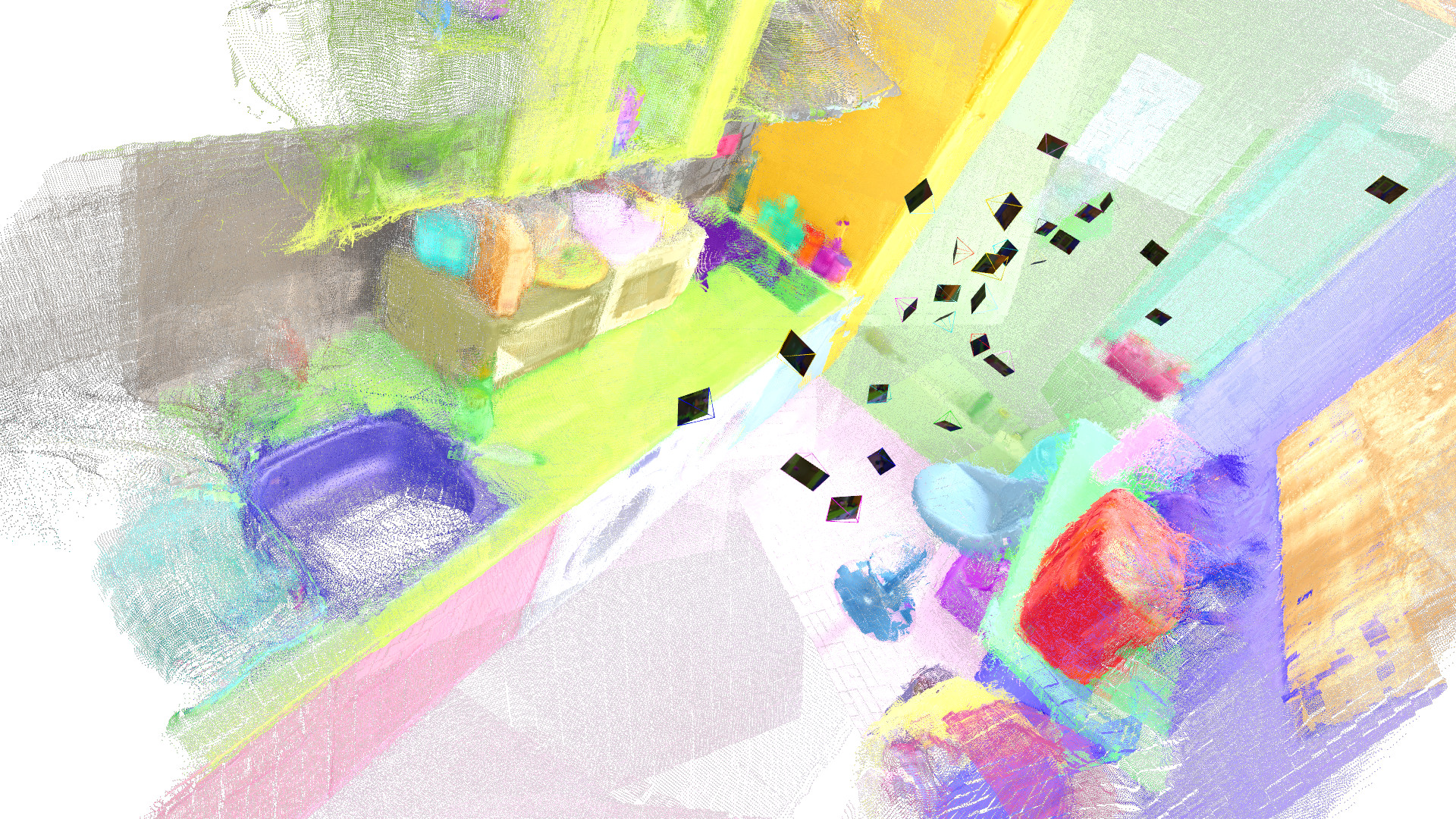}\\ \, \\
      \includegraphics[width=0.9\linewidth]{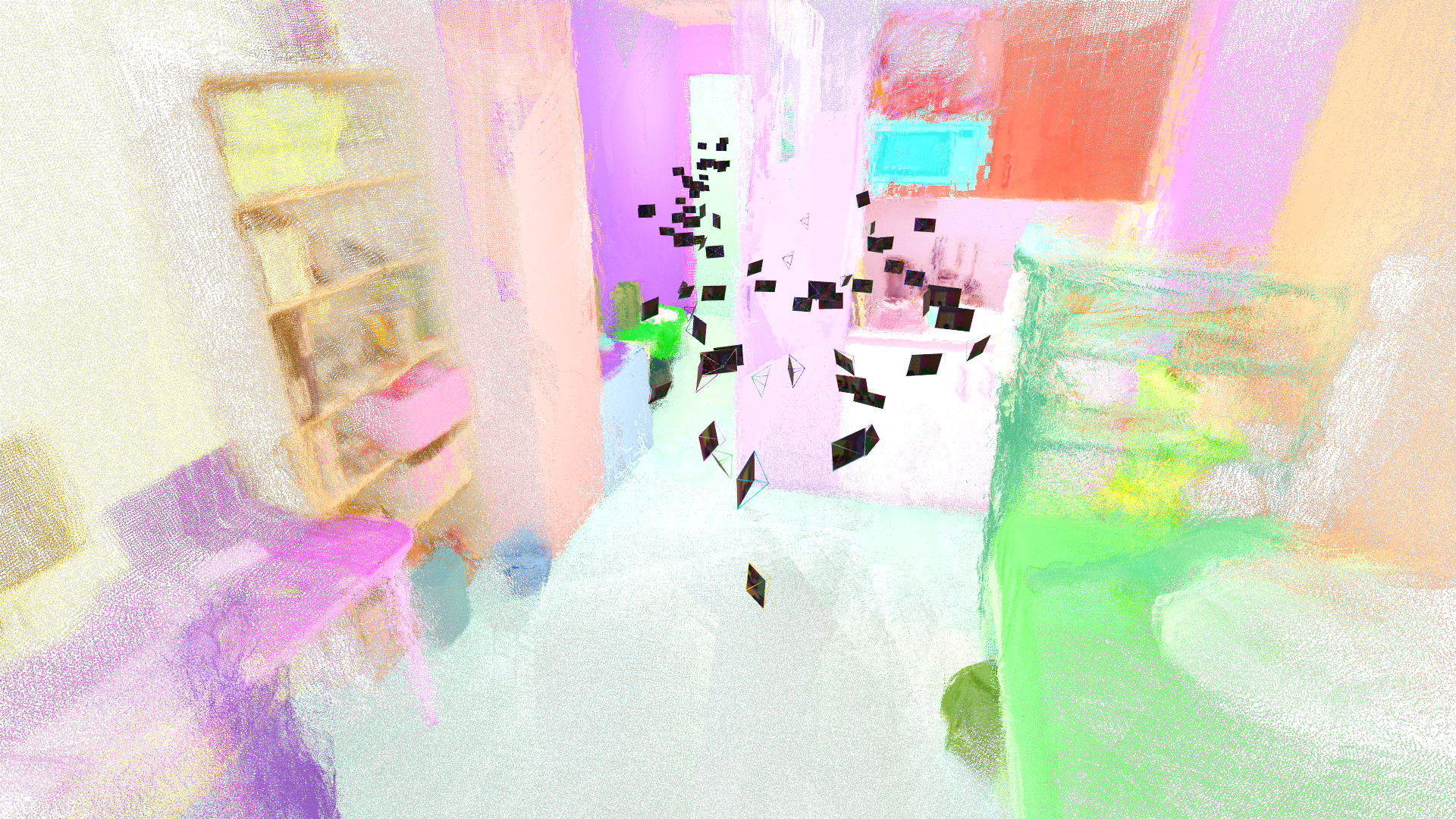}
      \\ \, \\
      \includegraphics[width=0.9\linewidth]{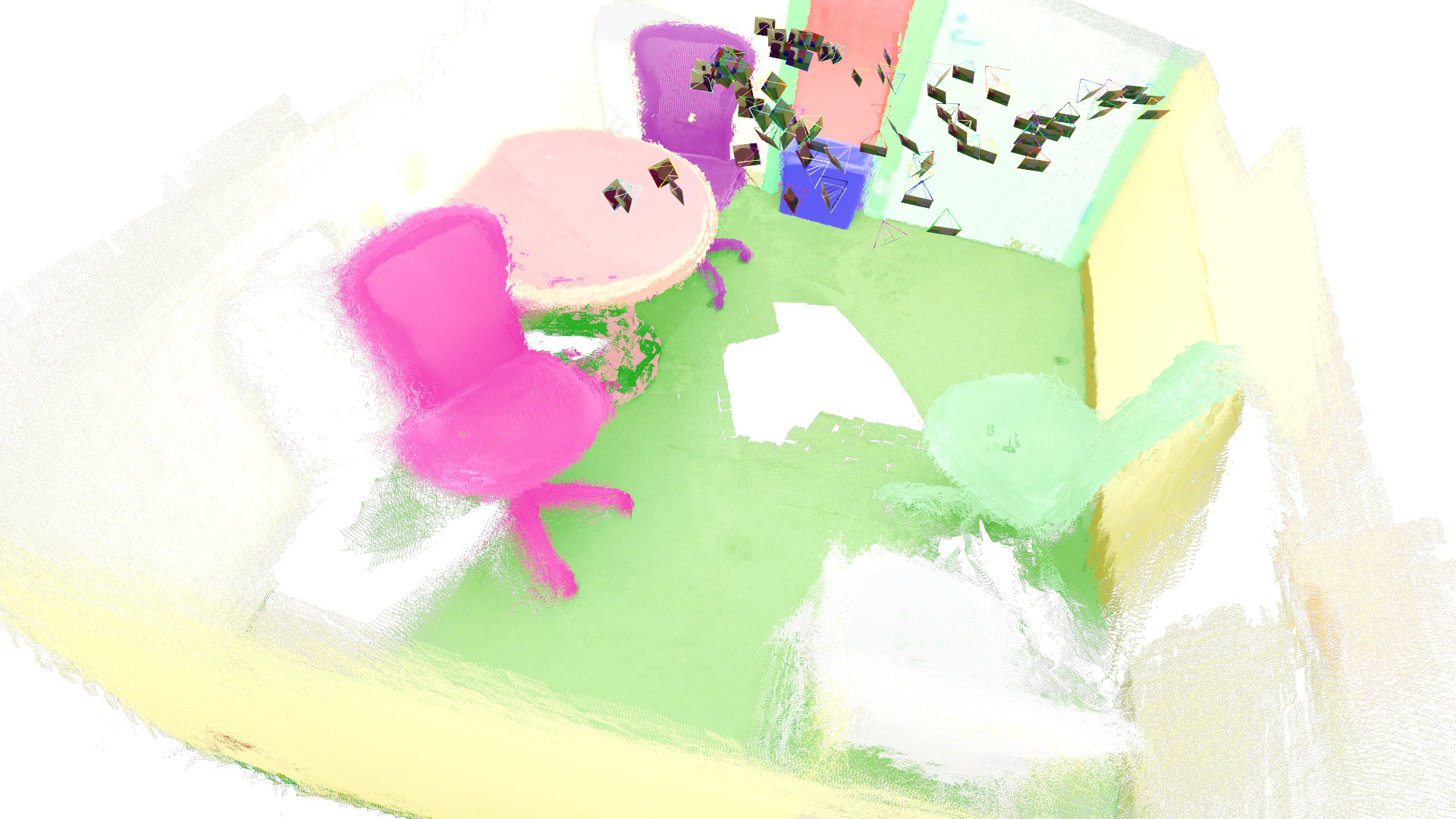}

    \caption{
        Further qualitative example of \panster{}'s segmented reconstruction.
        \panster{} outputs these labeled point-cloud in a single forward pass, without requiring any camera parameters nor test-time optimization.
    }
    \label{fig:qualitative}
\end{figure}

\section{Further qualitative results}

In \cref{fig:qualitative} we provide further qualitative results for direct panoptic prediction on a set of test images. In this case we simply color the predicted 3D points obtained with the 3D head with the unique "color"  corresponding to the object instance according to the selected "query" masks.

\section{\panster{} pseudo-code}

We provide pseudo-code of the \panster{} model architecture in Figures~\ref{code:p1}, \ref{code:p2}, \ref{code:p3} and \ref{code:p4}.

\begin{figure*}[th]
\begin{mypython}
def panst3r(frames, class_names):
    # frames: unordered collection of N frame RGB images [N, H=512, W=512, 3]
    # class_names: list of class names used in classification

    # 1.a Extract DINOv2 features (computed per-frame)
    dino_enc_feats = dinov2(frames)  # [N, T=HxW/(16x16), D_dino=1024]

    # 1.b Extract MUSt3R features (multi-view)
    must3r_enc_feats, must3r_dec_feats = must3r(frames) # [N, T, D_enc=1024], [N, T, D_dec=768]

    # 2. Concatenate features along channel dimension
    feats = concatenate(dino_enc_feats, must3r_enc_feats, must3r_dec_feats) # [N, T, D_enc+D_dec+D_dino]

    # 3. Prepare frame tokens and mask features
    frame_tokens = MLP(feats) # [N, T, D=768]
    frame_tokens = concatenate(frame_tokens) # [T*N, D]

    mask_feats = upscaler(feats) # [N, H/2, W/2, D_mask=256]

    # 4. Mask transformer
    # in_queries - set of learnable instance queries [Q=200, D]
    output_queries = mask_transformer(in_queries, frame_tokens) # [Q, D]

    # 5. Prediction heads
    out_masks, out_classes = prediction_heads(output_queries, mask_feats, class_names) # [Q, N, H/2, W/2], [Q, num_classes]

    # 6. Postprocessing w/ QUBO
    instance_mask, class_mask = QUBO(out_masks, out_classes) # [N, H/2, W/2], [N, H/2, W/2]
\end{mypython}
    \vspace{-.5cm}
    \caption{Pseudo-code of PanSt3R.}
    \label{code:p1}
\end{figure*}

\begin{figure*}[th]
\begin{mypython}
def mask_transformer(in_queries, frame_tokens):
    queries = in_queries
    for i in range(num_layers):
        # 1. Cross attention with frame tokens
        queries = masked_cross_attention(queries, frame_tokens)

        # 2. Self attention
        queries = self_attention(queries)

        # 3. Feedforward
        queries = feedforward(queries)

    return queries
\end{mypython}
    \vspace{-.5cm}
    \caption{Pseudo-code of PanSt3R: mask transformer architecture}
    \label{code:p2}
\end{figure*}

\begin{figure*}[th]
\begin{mypython}
def prediction_heads(output_queries, mask_features, class_names):
    # output_queries: refined queries, output of mask transformer [Q, D=768]
    # mask_features: upscaled feature maps [N, H/2, W/2, D_mask=256]
    # class_names: list of class names used in classification [C]

    class_embeddings = SigLIP_text(class_names) # [C, D]

    # 1. Class prediction (per-query class probability distribution)
    queries_cls = project(output_queries) # [Q, D] -> [Q, D]
    class_probs = cosine_similarity(queries_cls, class_embeddings) # [Q, C]

    # 2. Mask prediction (for each query a set of masks for all frames segmenting the same object instance)
    queries_mask = project(output_queries) # [Q, D] -> [Q, D_mask]
    mask_preds = dot_product(queries_mask, mask_features) # [Q, N, H/2, W/2]

    return mask_preds, class_probs
\end{mypython}
    \vspace{-.5cm}
    \caption{Pseudo-code of PanSt3R: prediction heads}
    \label{code:p3}
\end{figure*}

\begin{figure*}[th]
\begin{mypython}
def upscaler(feats):
    # Rearange patch tokens to 2D grid
    feats2d = rearange(feats) # [N, T=HxW/(16x16), D_enc+D_dec+D_dino=2816] -> [N, H/16, W/16, D_enc+D_dec+D_dino]

    for i in range(3):
        # Learnable upsampling, while decreasing the number of channels
        feats2d = upsample2x(feats2d) # [N, h, w, ...] -> [N, 2*h, 2*w, ...]

    return feats2d # [N, H, W, D_mask=256]
\end{mypython}
    \vspace{-.5cm}
    \caption{Pseudo-code of PanSt3R: upscaler architecture}
    \label{code:p4}
\end{figure*}

\end{document}